\documentclass[conference]{IEEEtran}
\IEEEoverridecommandlockouts

\usepackage{cite}
\usepackage{hyperref}
\usepackage{amsmath,amssymb,amsfonts}
\usepackage{listings}
\usepackage{algorithmic}
\usepackage{graphicx}
\usepackage{textcomp}
\usepackage{xcolor}
\usepackage{url}
\usepackage{verbatim}
\usepackage{hyperref}
\usepackage{amsmath}
\usepackage{pgfplots}
\usepackage{booktabs}
\usepackage{tikz}
\usepackage{caption}
\usepackage{enumitem}

\def\BibTeX{{\rm B\kern-.05em{\sc i\kern-.025em b}\kern-.08em
    T\kern-.1667em\lower.7ex\hbox{E}\kern-.125emX}}
    
    \definecolor{lightGray}{gray}{0.9}
\lstdefinelanguage{Prolog}{
  morekeywords={:-, ?, not},      
  sensitive=true,                 
  morecomment=[l]{\%},            
  morestring=[b]',                
}

\lstdefinelanguage{ASP}{
  morekeywords={:-, not},         
  sensitive=true,                 
  morecomment=[l]{\%},            
  morestring=[b]"                 
}

\begin{document}

\newcommand{\nb}[2]{
    \fbox{\bfseries\sffamily\scriptsize#1}
    {\sf\small\textcolor{red}{\textit{#2}}}
}

\newcommand{\nbc}[2]{
    \fbox{\bfseries\sffamily\scriptsize#1}
    {\sf\small\textcolor{blue}{\textit{#2}}}
}

\newcommand\ag[1]{\nb{AG}{#1}}
\newcommand\li[1]{\nbb{Roxana}{#1}}
\title{Ontology-guided multi-agent reasoning for explainable differential diagnosis}

\author{\IEEEauthorblockN{Roxana Ramona Szomiu, George Ursu, Adrian Groza}

\IEEEauthorblockA{
        \textit{Artificial Intelligence Research Institute AIRi@UTCN} \\
        \textit{Department of Computer Science, Technical University of Cluj-Napoca} \\
        Cluj-Napoca, Romania \\
        Email: Roxana.Szomiu@cs.utcluj.ro, Ursu.Da.George@student.utcluj.ro, Adrian.Groza@cs.utcluj.ro
    }
    }
    
\maketitle

\begin{abstract}
Effective differential diagnosis depends on well-structured and unambiguous patient case descriptions, since incomplete or ambiguous clinical information readily leads to diagnostic errors and inconsistent reasoning. Large Language Models (LLMs) have recently shown promising capabilities in medical question answering and clinical reasoning; however, even when achieving competitive performance on medical benchmarks, they often rely on pattern recognition learned from training data rather than grounded clinical evidence, and can generate plausible yet unsupported diagnostic conclusions. Their reasoning remains largely opaque, making it difficult for clinicians to assess the evidence and rationale behind a given recommendation.

To address this gap, we propose an ontology-guided, multi-agent differential diagnosis framework that replicates the iterative reasoning process of physicians. The framework grounds diagnostic inference in a structured, expert-curated medical knowledge base, enabling interpretation of symptoms, clinical findings, or disease relationships, while decomposing diagnosis into explicit, auditable reasoning steps. 
This ontology-based grounding constrains inference to evidence-backed associations, reducing hallucination and improving diagnostic transparency, traceability, and consistency. The system then identifies the most discriminative missing symptoms and generates targeted follow-up questions to reduce uncertainty through a process that mirrors to clinical inquiry. Together, these components improve the reliability, explainability, and diagnostic accuracy of AI-assisted differential diagnosis. The source code is publicly \href{https://github.com/georgeursu15/MedMAS-DDx}{available}.
\end{abstract}


\section{Introduction}
Differential diagnosis is an iterative process used to distinguish between two or more diseases with similar symptoms. 
Initial patient notes are often incomplete, making reliable diagnostic decision-making difficult without further investigation.
\footnotetext[1]{Roxana Ramona Szomiu was supported by an Adobe Romania grant. Adrian Groza was supported by the Romanian Hub for Artificial Intelligence (HRIA), funded through the Smart Growth, Digitization and Financial Instruments Program (MySMIS No. 334906). The authors gratefully acknowledge the administrative and
technical support provided by the Technical University of Cluj-Napoca during the development of this research.}
Early evaluations of LLMs in healthcare formulated diagnostic reasoning as a \textit{single-turn question-answering task}, assuming that all the information required for diagnosis and treatment selection was available at the time of prediction \cite{app11146421}. 

Transitioning from static QA systems \cite{pal2022medmcqalargescalemultisubject} to interactive, multi-turn diagnostic dialogues is crucial for clinical decision-making.
Physicians iteratively collect information through targeted questions to reduce uncertainty, highlighting diagnosis as an interactive process rather than a single-step inference.

Motivated by the potential of LLMs to support differential diagnosis and clinical decision-making \cite{mcduff2023accuratedifferentialdiagnosislarge}, we propose an ontology-based multi-agent system (MAS) that replicates iterative clinical reasoning and assists clinicians in diagnosis.

Our neuro-symbolic framework integrates LLM-based language understanding with ontology-driven symbolic reasoning. Organized into two modules, it employs a task-oriented multi-agent architecture to interpret clinical narratives, extract symptoms, and manage patient interactions.

The symbolic reasoning module, grounded in medical ontologies (SYMP~\cite{Schriml2010-cc}, DOID~\cite{Baron2026DOKB},
and HPO \cite{Robinson2008-ov}), performs transparent disease scoring and differential diagnosis.
The module queries the medical knowledge graph (KG) to identify the most informative symptoms that distinguish among competing diagnoses. 
Based on this analysis, the dialogue-generator agent autonomously formulates targeted Yes/No follow-up questions that efficiently reduce diagnostic uncertainty, and maximize diagnostic information gain. Rather than relying on a single prediction from a fixed set of symptoms, the framework incrementally refines the differential diagnosis as new clinical evidence becomes available. 

By combining LLMs with ontology reasoning, the proposed MAS supports an adaptive and explainable diagnostic workflow. 
Explainability is achieved through traceable decision-making: mathematical scoring functions rank diagnoses based on extracted evidence and ontology relationships. The system links each refinement with the contributing symptoms and scores, allowing clinicians to transparently interpret the reasoning behind the diagnosis.
This enables a transparent, clinically meaningful interpretation of the diagnosis and its reasoning by clinicians.

In summary, our main contributions are as follows:
\begin{itemize}[topsep=0pt, itemsep=0pt, parsep=0pt, partopsep=0pt]
    \item ontology-guided multi-agent framework: We propose a neuro-symbolic framework that separates language understanding from clinical reasoning and grounds diagnostic scoring in expert-curated ontologies, reducing LLM hallucinations.
    \item interactive clinical reasoning: The system generates targeted follow-up questions for missing discriminative symptoms, reducing diagnostic uncertainty and compensating for up to 30\% loss in initial symptom disclosure.
    \item traceable explainability: 
    Mathematical scoring functions rank diagnoses at each stage, enabling traceable diagnostic refinement.
\end{itemize}

\section{Related work}

Singhal et al.~\cite{Singhal2025-sn} introduced Med-PaLM2, a medical LLM that achieved expert-level performance in several clinical question-answering benchmarks, including 86.5\% accuracy on MedQA, representing a more than 19\% improvement over Med-PaLM \cite{singhal2022largelanguagemodelsencode}. 
However, single-turn question-answering has a key limitation: even fully specified symptoms rarely finalize a diagnosis. In practice, doctors must actively gather missing information through targeted follow-up questions.

Muntean et al.~\cite{diagnostics14141468} demonstrate that Med-PaLM2 can provide accurate and clinically appropriate responses for patient education tasks, to ophthalmology-related questions. However, the interaction is limited to a single-turn exchange, which prevents the model from clarifying ambiguities or refining differential diagnoses through iterative dialogue.
This limitation motivates a shift from static QA benchmarks to interactive diagnosis, where diagnoses are progressively refined through iterative doctor-patient questioning.

The Q4Dx~\cite{Werthaim2026-vx} benchmark evaluates the strategic inquiry capability of diagnostic systems. It assumes real encounters begin with incomplete, noisy descriptions, requiring doctors to recover missing information through adaptive, context-dependent questioning. 
Synthetic patient notes are generated from the SymbiPredict dataset~\cite{Tucker2024-bu} at various completeness levels (100\%, 80\%, 50\%). GPT-4.1 and GPT-4o-mini agents simulate patient-physician interactions to assess diagnostic uncertainty reduction over multiple turns.
The study reveals that while partial information significantly reduces single-shot diagnostic performance, interactive questioning can recover much of this loss, underscoring the importance of dialogue-based reasoning in clinical diagnosis. 
Baseline accuracy declines from 51\% to 44\% and 37\% as symptom disclosure decreases from 100\% to 80\% and 50\%, respectively.

Recent work on interactive clinical reasoning introduces complementary benchmarks and frameworks that together advance multi-turn diagnostic modeling. 
MedKGI  ~\cite{wang2026medkgiiterativedifferentialdiagnosis} improves diagnostic dialogue by grounding reasoning in medical knowledge graphs, selecting questions based on information gain, and maintaining structured state tracking, leading to more efficient and coherent interactions. MedPI ~\cite{v2025medpievaluatingaisystems} evaluates flagship models across 7,100 simulated conversations, showing that despite strong static benchmark performance, LLMs struggle with full patient-clinician dialogues. Similarly, MINT ~\cite{fang2026benchmarkingmultiturnmedicaldiagnosis} highlights issues like premature diagnostic commitment, sensitivity to misleading evidence, and limited self-correction. They show that delayed questioning and structured evidence presentation significantly improve accuracy by up to 62,6\% . Overall, these studies demonstrate that effective clinical diagnosis requires grounded, multi-turn, and temporally aware reasoning rather than single-step prediction.

Gond et al.~\cite{gong2026meddialogrubricscomprehensivebenchmarkevaluation} introduce MedDialogRubrics, a benchmark of 5,200 synthetic patient cases and over 60,000 expert-refined rubrics for multi-turn clinical consultations, and show that even strong models (e.g., Gemini-2.5-Pro) miss nearly half of the must-ask questions. 
Their analysis shows that diagnostic performance peaks at approximately 10-11 turns and declines with longer interactions.
These findings motivate limiting the interviewer agent to 11 follow-up questions, which maximizes accuracy while avoiding reduced diagnostic alignment and unnecessary consultation length.

MedAgents \cite{tang2024medagentslargelanguagemodels} is one of the first multi-agent collaboration frameworks designed specifically for the medical domain. Evaluated on nine  multiple-choice datasets, MedAgents reaches an average accuracy of 86.7\% with GPT-4, slightly above a stronger few-shot CoT (Chain-of-Thought) baseline despite using no in-context examples.

\section{Methodology}

MedMAS-DDX is a multi-agent system (MAS) that treats differential diagnosis as a multi-step process, not a single inference. It translates patient's free-text complaint to a ranked, explained list of probable diseases through a sequence of transformations: unstructured language becomes structured symptoms, symptoms become candidate diseases, and candidate diseases are organized into an initial ranking representing the differential diagnosis. That ranking is then sharpened by a short exchange of questions, and the answers are fed back into the scoring to produce the final diagnosis.

Two principles govern every step. First, responsibility is split by reliability: language models handle only natural language understanding and generation, while all ranking-related computations are done by deterministic code, ensuring that diagnosis does not depend on model-based calculations.
Second, all medical knowledge is stored in an external curated source and is explicitly queried during the process, rather than being learned from the model’s parameters.

\begin{figure*}
    \centering    \includegraphics[width=\textwidth]{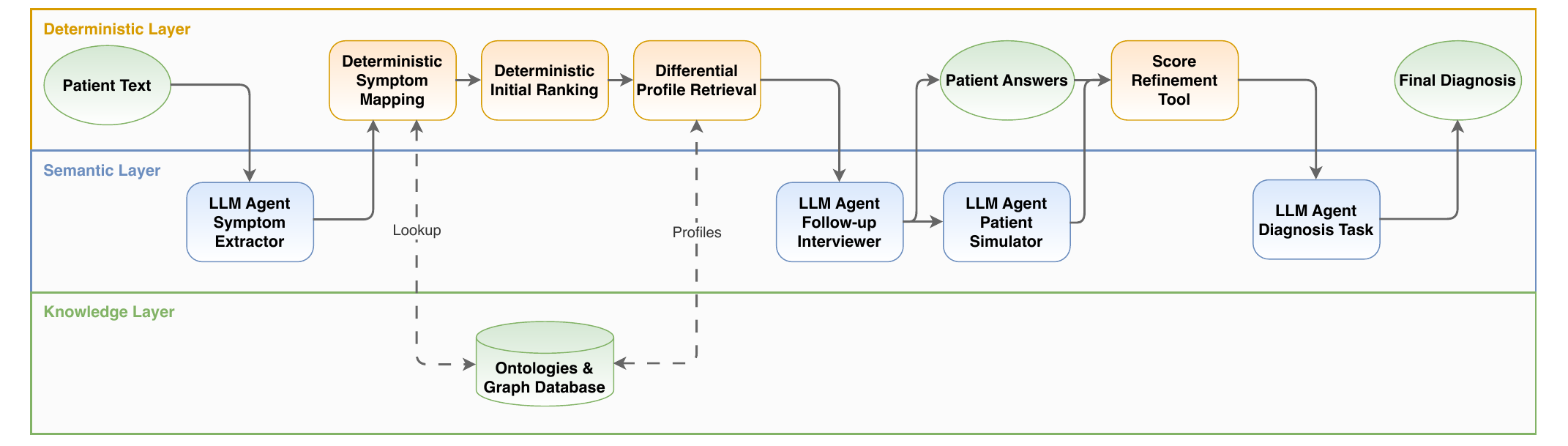}
    \caption{System components and their interactions.}
    \label{fig:architecture-diagram}
\end{figure*}

MedMAS-DDX realizes the preceding principles as a neuro-symbolic pipeline in which a small set of LLM-based agents is embedded within, and coordinated by, a deterministic control layer (Figure~\ref{fig:architecture-diagram}).
The multi-agent system is implemented using the CrewAI  \footnote{\url{https://crewai.com/}}
framework, which manages agent definition and execution. Each agent is designed as a distinct role-goal-backstory persona operating based on the same LLM (via LiteLLM \footnote{\url{https://www.litellm.ai/}}), with specialized, non-interchangeable functions.

The system consists of four agents: 
(i) a symptom extraction agent (\textit{symptom\_extractor}), 
(ii) an interviewing agent (\textit{followup\_interviewer}), 
(iii) an explanatory agent (\textit{diagnosis\_refiner}), 
(iv) a patient agent (\textit{patient\_simulator}) used only to drive automated evaluation and never contributes to a live diagnosis. 
The first three agents participate in the diagnostic-making process, while the last agent simulates the patient.

The MAS is characterized by high control over execution, predictable processing paths, making it well-suited for strict business logic. 
In this architecture, agents communicate indirectly through deterministic Python-based coordination.
This control layer validates and transforms intermediate representations, ensuring structured, code-mediated interactions rather than unconstrained agent-to-agent dialogue.
The diagnostic-making pipeline allocates specific tasks to three active agents during a live diagnosis, while a fourth persona is reserved exclusively for methodological validation.

\subsection{Symptom extractor agent} 

The symptom extractor agent, an LLM-based component without external tool access, constitutes the first step of the pipeline by parsing the patient’s initial, natural-language clinical note - often incomplete and informal - and transforming it into a structured intermediate representation of symptoms. For each identified symptom, it generates a canonical medical term, alternative phrasings, and optional contextual details such as duration or location, while keeping descriptive information separate from standardized terminology to ensure consistency and support ontology matching in the next step.

Our solution follows the methodology of extracting information from unstructured clinical notes and map them to corresponding ontology terms \cite{STINTON2023100007}, ~\cite{info:doi/10.2196/73233}.
According to this, in the second phase of the pipeline, the linking symptoms to diseases process occurs in two stages-\textit{resolution and retrieval}-and constitutes the first interaction with external knowledge. In the\textit{ resolution stage}, each symptom is deterministically mapped to a standard medical concept through an ordered matching process that first attempts the canonical term, then alternative phrasings, and finally a secondary synonym vocabulary. 
Unresolved symptoms are retained for traceability, but excluded from further processing to avoid affecting the following inference steps.

In the \textit{retrieval stage}, resolved concepts query a knowledge graph of expert-defined symptom-disease relationships via a consolidated lookup. Diseases enter the candidate set only if explicitly linked to at least one patient symptom, ensuring grounding in verifiable knowledge.

All medical knowledge is stored in a GraphDB \footnote{\url{https://graphdb.ontotext.com/}} triple store and queried exclusively via SPARQL, ensuring no disease-symptom relations are drawn from LLM hallucinations.

The knowledge base spans two repositories: one for the Symptom (SYMP) and Human Disease (DOID) Ontologies, and another for the Human Phenotype Ontology (HPO), acting as a synonym bridge during resolution.
Since queries are executed in Python rather than by LLM-based agents, knowledge access remains uniform, inspectable, and model-independent.
 
The retrieved candidate diseases and their associated symptoms advance to the third phase: ranking. Here, diseases are deterministically ranked based on symptom matches, assigning higher weights to rarer, more informative symptoms to produce the initial differential diagnosis.

\begin{equation}\label{eq:1} 
    specificity\_weight(s)=
    \frac{1}{\log_{2}\!\left(1+N_d(s)\right)}
\end{equation}
Eq. \ref{eq:1} encodes a key idea that rare symptoms are more informative than common ones. Let $N_d(s)$ denote the number of diseases associated with symptom \textit{s }in the KG, and the logarithm reduces the impact of very large disease counts.
If a symptom \textit{s} appears in many diseases, then \textit{$N_d(s)$ } is large, the denominator is large
so weight becomes small (low specificity).
If a symptom \textit{s} appears in few diseases, then
\textit{$N_d(s)$} is small, the denominator is small so weight becomes large (high specificity).

For each candidate disease \textit{d}, a raw diagnostic score is computed by summing the diagnostic weights of all matched symptoms:

\begin{equation}\label{eq:2} 
raw\_score(d)
=
\sum_{s \in \text{matched}(d)}
specificity\_weight(s)
\end{equation}

A \textit{coverage factor} measures the proportion of the patient's mapped symptoms explained by each candidate disease \textit{d}:

\begin{equation}\label{eq:3} 
    coverage(d)
    =
    \frac{match\_count(d)}
    {total\_mapped\_symptoms}.
\end{equation}

\textit{match\_count(d)} denotes symptoms matched by disease \textit{d}, while \textit{total\_mapped\_symptoms} denotes all successfully mapped symptoms.
The \textit{coverage} factor rewards diseases explaining a larger fraction of reported symptoms, complementing the informativeness-based raw score.

The raw score \ref{eq:2} and the coverage factor \ref{eq:3} are then combined into a final \textit{ composite score}:
\begin{equation}\label{eq:4} 
    comp\_score(d)    = raw\_score(d)\times\bigl(0.5 + 0.5\times coverage(d)\bigr)
\end{equation}
The \textit{ composite score }combines two complementary aspects of diagnostic quality: the strength of supporting evidence and the completeness of symptom coverage. Here,\textit{ raw\_score(d) } measures the total diagnostic evidence for disease \textit{d}, computed from the weighted contributions of matched symptoms. In contrast, \textit{coverage(d)} measures how much of the patient’s symptom set is explained by the disease \textit{d}.

The term $(0.5 + 0.5 \cdot \text{coverage}(d))$ is a bounded scaling factor in [0.5,1] that adjusts rather than dominates the raw score. It ensures that coverage acts as a soft adjustment to the diagnostic evidence: when (\text{coverage}(d)=0), the score is reduced but not eliminated (scale (=0.5)), while full coverage preserves the raw score (scale (=1)).
This design prevents excessive penalization of low-coverage candidates while maintaining stable ranking behavior and balancing evidence strength with explanatory completeness.

An inclusive threshold then keeps every candidate that is plausibly relevant: those that reach a small fraction of the leading score, or that match at least two symptoms. 
The threshold is set to a low value on purpose because a candidate discarded here cannot be recovered later.
At this stage, a single remaining candidate is returned directly; otherwise, the process proceeds to questioning.
 
\subsection{Follow-up interviewer agent} 

When multiple candidate diseases remain, the system enters a differentiation phase. The characteristic symptom profiles of the leading candidates are retrieved from the knowledge source and combined with the current ranking and reported symptoms to form an enriched state.  The follow-up interviewer agent is responsible for resolving ambiguities and collecting missing clinical information.
The interviewer agent filters non-discriminative and previously reported symptoms, selects patient-observable ones, and generates targeted Yes/No questions to refine the differential diagnosis.

At this stage, we have a human-in-the-loop interaction point in the system. After the system generates follow-up questions, the process stops temporarily to allow the patient (or user) to respond. The responses are constrained to a simple, structured format-“yes”, “no”, or “unsure”-which makes the input easy to process deterministically and avoids ambiguity in following reasoning steps.
The mechanism that collects the answers is kept separate from the diagnostic logic, so the same pipeline runs unchanged whether answers come from a command line, a web interface, or an automated non-interactive source. 
This step turns a one-shot classification into a consultation. The system has formed the initial hypotheses and identified the most informative missing facts, and it obtains exactly those before concluding, much as a clinician narrows the possibilities through targeted questioning.

\subsection{Patient simulator agent} 
To ensure rigorous and automated pipeline evaluation, a fourth persona is employed exclusively as a testing mechanism. The patient agent receives a fully specified case description and generates responses to the questions posed by the \textit{followup\_interviewer}, based only on the information contained in the current case. 
This agent operates entirely outside the production environment, with no external tool access and no involvement in live diagnostic sessions, serving exclusively for system validation purposes.

\subsection{Diagnosis refiner agent} 

The diagnosis refiner agent serves as the final analytical component of the live diagnostic process, synthesizing all iteratively collected findings into a coherent medical conclusion. 
It uses confirmed symptoms and disease-symptom relations retrieved from GraphDB to produce a final diagnosis with a clear, evidence-based explanation.

The patient agent's answers are reintroduced into the pipeline and used to update the differential diagnosis through deterministic scoring. For each queried symptom \textit{ s}, a \textit{discriminative weight} is computed as:

{\small
\begin{equation}\label{eq:5} 
        discriminative\_weight(s) = 1 - \frac{diseases\_with\_symptom(s)}{n_{profiled}}
\end{equation}
}
The \textit{diseases\_with\_symptom(s)} denotes the number of profiled candidate diseases that present symptom\textit{ s}, and $n_profiled$ is the total number of profiled candidate diseases. This weight quantifies the ability of a symptom to distinguish among the remaining candidates: symptoms shared by many diseases receive lower weights, whereas symptoms present in only a few candidates receive higher weights. Consequently, responses to highly discriminative symptoms have a greater influence on the final ranking, while common symptoms contribute less because they provide limited additional diagnostic information.

This approach penalizes common features, similar to the established IDF (Inverse Document Frequency) principle in information retrieval. However, rather than adopting the standard logarithmic scaling of IDF 
($\log(n_{profiled}/diseases\_with\_symptom(s)) $
), Eq. \ref{eq:5} utilizes a linear complement.
This design choice is tailored to the specific constraints of the diagnostic pipeline and bounds discriminative weights to [0, 1), ensuring mathematical stability in subsequent multiplicative score updates.
Moreover, while logarithmic smoothing is necessary for massive document corpora, a linear formulation provides a more distinct and clinically intuitive separation of symptom specificity when evaluating the small and dynamic pools of candidate diseases typical in differential diagnosis.

For each disease, only the relevant questions are considered, that is, those whose symptom lies in its profile. The \textit{discriminative\_weights} of the confirmed symptoms and of the denied symptoms are summed and normalised by the number of relevant questions into a \textit{support term} and a \textit{conflict term}. These adjust thfe prior \textit{ composite score} \ref{eq:4} through a bounded multiplicative update:
{\small
\begin{multline}
    final\_score(d) = composite\_score(d) \times {} \\
    \bigl(1 + \alpha\,support\_norm(d) - \beta\,conflict\_norm(d)\bigr)
\end{multline}
}

where confirmations increase the score and denials decrease it according to their discriminative weights.

Both the \textit{support\_term} and the \textit{conflict\_term} are normalized to the $[0, 1]$ interval. The hyperparameters $\alpha$ and $\beta$ 
control the relative contribution of supporting and
conflicting evidence, respectively. 
 Given that supporting evidence has substantially greater diagnostic significance than conflicting evidence in our framework, we established $\alpha$ = 0.6 and $\beta$ = 0.2. These values were derived by empirical hyperparameter tuning on a validation subset. We systematically evaluated multiple parameter configurations  
 and the combination of $\alpha=0.6$ and $\beta=0.2$ yielded optimal diagnostic accuracy and rank stability. 
 This configuration reflects a fundamental diagnostic principle: the presence of a
highly specific symptom is typically stronger positive evidence
for a disease than its absence is negative evidence, as symptoms may be unobserved by the patient, or simply delayed in their onset. 

 In contrast, higher conflict-penalty weights ($\beta = 0.4$ and $\beta = 0.6$) overly penalized plausible diagnostic candidates with temporarily unobserved symptoms, leading to their premature elimination from the differential.

By limiting $\beta$ relative to $\alpha$, the model keeps the adjustment factor bounded, allowing follow-up responses to refine the initial diagnostic ranking without replacing it or artificially amplifying previously established evidence.
The candidates are re-ranked and the most influential answer is identified.

When the final diagnosis is determined, the language model returns to explain. It receives the computed scores and states, for each disease, why it rose or fell in the ranking. 
Each disease's confidence is determined by the score margins instead of being asserted: high for a clear leader, cautious when the margin is narrow. The output is therefore a re-ranked differential in which every disease carries a calibrated confidence and an transparent, evidence-based explanation.

\section{Experiments and Results}

This section evaluates MedMAS-DDx based on its diagnostic trajectory -its ability to acquire missing information through follow-up questions and iteratively refine the differential diagnosis - rather than as a static classifier.

The active agents access LLMs via LiteLLM (remote API or self-hosted vLLM). All deterministic scoring and ranking processes run locally, independent of the LLM host.
Section~\ref{sec:qualitative-case-study} traces one evaluation case end to end to illustrate the mechanism concretely.

\subsection{Evaluation of framework and experimental setup}
The framework is evaluated on the Q4DX benchmark \cite{Werthaim2026-vx} using \textit{PatientCases.csv} (99 cases, 29 diagnoses) at two symptom-completeness levels: 80\% and 50\%. The 100\% setting is excluded because no symptoms remain for follow-up. This yields $N = 198$: 99 samples at $80\%$, 99 samples at $50\%$.

Follow-up questions are answered by a deterministic simulated patient agent with access to the full symptom profile (case 100\%), providing an optimistic estimate of interactive information recovery rather than real-patient performance.
Diagnostic correctness is measured using SapBERT cosine similarity with the reference label (target diagnosis), using a 0.6 threshold to tolerate differences in medical terminology.

\subsection{Diagnostic accuracy and interactive refinement}

\begin{figure}[ht]
  \centering
  \begin{tikzpicture}[scale=0.8]
    \begin{axis}[
        xbar stacked,
        bar width=17pt,
        width=1.1\linewidth,
        height=2.5cm,
        xmin=0, xmax=100,
        xlabel={Proportion of Extracted Symptoms (\%)},
        ytick=\empty,
        point meta=explicit symbolic, 
        nodes near coords, 
        nodes near coords align={center},
        every node near coord/.append style={font=\footnotesize, xshift=-10.5pt},
        legend style={at={(0.5,1.03)}, anchor=south, legend columns=-1, column sep=1ex, font=\small},
        area legend,
        enlarge y limits=0.5,
    ]
      \addplot coordinates {(77.3,0) [77.3\%]}; 
      \addplot coordinates {(14.0,0) [14.0\%]}; 
      \addplot coordinates {(8.7,0) [8.7\%]};  
      
      \legend{Primary (SYMP), Fallback (HPO), Unmapped}
    \end{axis}
  \end{tikzpicture}
  \caption{Distribution of symptom mapping resolution} 
  \label{fig:symptoms_mapping}
\end{figure}
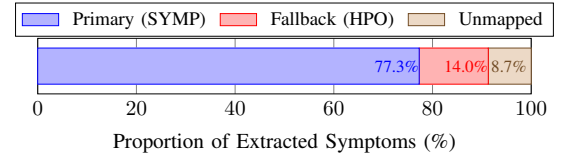

Symptom mapping was verified before assessing accuracy: 91.3\% of 1,243 symptoms were mapped to ontology concepts. As Figure~\ref{fig:symptoms_mapping} illustrates, direct matching to the primary SYMP ontology covered 77.3\% of symptoms. To minimize information loss, a sequential fallback strategy queried unresolved symptoms against the HPO ontology, recovering an additional 14.0\% (e.g., “passage of gases” $\rightarrow$ flatulence) and capturing expressions absent from SYMP. 

The remaining 8.7\% unmapped symptoms primarily result from the LLM extracting overly complex descriptions lacking direct ontology equivalents. Furthermore, the fallback mechanism's lexical tolerance occasionally introduces semantic errors (e.g., mapping “unsteady” to “hemodynamic instability”).

Interactive refinement (Phase 2) consistently improves the zero-shot differential (Phase 1). Of the 42 correct top-1 predictions in Phase 2, only 6 were exact string matches, validating our use of semantic evaluation (SapBERT) over strict string matching.

At the aggregate level, interactive refinement (Phase 2 follow-up) improves on the zero-shot differential (Phase 1) across every metric, including the mean semantic similarity of the top-ranked prediction computed with SapBERT semantic evaluation at the $0.6$ threshold.(Table~\ref{tab:global-accuracy}).

\begin{table}[ht]
\centering
\caption{Global diagnostic accuracy: Phase~1 and Phase~2 }
\label{tab:global-accuracy}
\small
\begin{tabular}{l c c}
\toprule
\textbf{Metric} & \textbf{Phase 1} & \textbf{Phase 2} \\
\midrule
Top-1 accuracy & $18.7\%$ ($37/198$) & $21.2\%$ ($42/198$) \\
Top-3 accuracy & $25.8\%$ ($51/198$) & $27.3\%$ ($54/198$) \\
Mean similarity (top-1) & $0.449$ & $0.466$ \\
Mean similarity (top-3) & $0.523$ & $0.537$ \\
\bottomrule
\end{tabular}
\end{table}

Table~\ref{tab:mean-similarity} presents an analysis of the mean semantic similarity per disclosure variant computed with SapBERT, identifying an important compensatory mechanism to our multi-agent architecture. By engaging in targeted follow-up questions, 
the system successfully recovers diagnostic confidence, matching the zero-shot baseline of cases with significantly richer initial narratives. This finding demonstrates that the interactive dialogue actively bridges the clinical information gap, effectively compensating for a 30\% loss in initial symptom disclosure.

\begin{table}[ht]
\centering
\caption{Mean semantic similarity: Phase~1 (initial differential) and Phase~2 (after follow-up)}
\label{tab:mean-similarity}
\small
\begin{tabular}{l c c}
\toprule
\textbf{Completeness Variant} & \textbf{Phase 1} & \textbf{Phase 2} \\
\midrule
$80\%$ Completeness & $0.4589$ & $0.4792$ \\
$50\%$ Completeness & $0.4394$ & $0.4527$ \\
\bottomrule
\end{tabular}
\end{table}

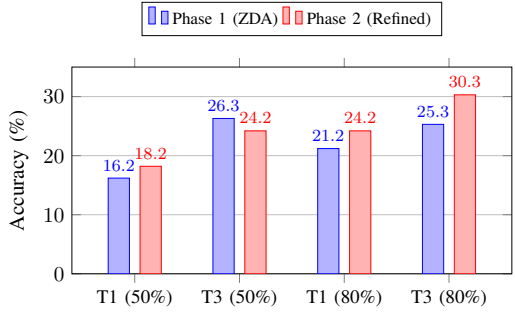
\begin{figure}[ht]
  \centering
  \begin{tikzpicture}[scale=0.8]
    \begin{axis}[
        ybar=5pt,
        bar width=10pt,
        width=\linewidth,
        height=5cm,
        enlarge x limits=0.20,
        ymin=0, ymax=35,
        ylabel={Accuracy (\%)},
        ylabel near ticks,
        symbolic x coords={T1 (50\%), T3 (50\%), T1 (80\%), T3 (80\%)},
        xtick=data,
        x tick label style={align=center, font=\small},
        nodes near coords,
        nodes near coords style={font=\footnotesize},
        legend style={at={(0.5,1.15)}, anchor=south, legend columns=-1, font=\footnotesize},
        ymajorgrids,
    ]
      \addplot coordinates {(T1 (50\%),16.2) (T3 (50\%),26.3) (T1 (80\%),21.2) (T3 (80\%),25.3)};
      \addplot coordinates {(T1 (50\%),18.2) (T3 (50\%),24.2) (T1 (80\%),24.2) (T3 (80\%),30.3)};
      \legend{Phase 1 (ZDA), Phase 2 (Refined)}
    \end{axis}
  \end{tikzpicture}
  \caption{Top-1 and Top-3 accuracy comparison between the initial Phase 1 and the interactively refined Phase 2.}
  \label{fig:accuracy_phases}
\end{figure}

Figure~\ref{fig:accuracy_phases} summarizes the results and shows, that refinement is not uniformly advantageous: performance improves from phase 1 to phase 2 in all cases, except for a slight decrease in Top-3 accuracy at $50\%$ completeness.
We use the Zero-shot Diagnostic Accuracy (ZDA) metric from the benchmark ~\cite{Werthaim2026-vx} for phase 1. This asymmetry, together with Case \#96 below, is presented in the Discussions section.

\subsection{Qualitative case study: Recovering the diagnosis of Hepatitis A (Case \#96)}
\label{sec:qualitative-case-study}

To illustrate how a single follow-up answer can reorder a competitive differential, we trace Case \#96 (\emph{Hepatitis~A}, $80\%$ completeness) end to end. Its canonical profile includes ten symptoms; the $80\%$ narrative omits \emph{yellowish skin} and \emph{yellowing of eyes} - i.e., jaundice, the most discriminative sign for hepatitis. The clinical narrative given to the pipeline at this completeness level reads as follows:

\begin{quote}
\emph{A 32-year-old male presented to the clinic with troubling symptoms, including persistent \textbf{joint pain} that has worsened over the past two weeks. He reported episodes of \textbf{vomiting}, \textbf{nausea}, and a noticeable \textbf{loss of appetite}. Moreover, he has been suffering from \textbf{abdominal pain}, accompanied by \textbf{diarrhea}. He also noted a \textbf{mild fever} and \textbf{muscle pain} that has hindered his daily activities, which has led to increased concern regarding his health status.}
\end{quote}

\paragraph{Phase 1a} The \textit{symptom\_extractor} maps all eight symptoms present in the narrative to SYMP concepts ($8/8$); \textit{jaundice }is simply absent from the text, not lost in mapping.

\paragraph{Phase 1b} Without the \textit{"jaundice"} symptom, \emph{"gastrointestinal anthrax"} and \emph{"oropharyngeal anthrax"} lead with a \textit{composite score} of $0.922$ (Eq.~\eqref{eq:4}), driven by the high \textit{specificity weight} of \emph{"severe diarrhea"} (Eq.~\eqref{eq:1}). Four \textit{"hepatitis"} subtypes (A, B, C, E) tie just below at $0.861$. The top prediction's SapBERT similarity to \emph{"Hepatitis A"} is only $0.340$ - an incorrect Top-1. Although the correct label is tied for third place, the prediction is still considered correct under the Top-3 metric.

\paragraph{Phase 1c} The \textit{followup\_interviewer} asks eleven targeted Yes/No questions (about ascites, acholic stool, dark urine, jaundice, liver inflammation, hematemesis, hematochezia, dyspnea, backache, headache, cough); the simulated patient confirms only \emph{"jaundice"} and denies the others.

\paragraph{Phase 2.} Confirming \textit{"jaundice"} (Eq.~\eqref{eq:5}) raises \textit{Hepatitis A, B, and C} to $0.872$ despite three denials; \textit{"Hepatitis E"} gains less, since an added denial on \emph{dark urine} drops it to $0.838$. Both \textit{"anthrax"} candidates receive no support and four denials, falling to $0.781$, below all four \textit{"hepatitis"} subtypes.

\paragraph{Outcome} \textit{"Hepatitis A"} becomes the top prediction, with similarity rising from $0.340$ to a perfect $1.0$ - the largest jump in the entire batch. A single confirmed symptom, combined with tie-aware scoring at the group level, was enough to resolve the ambiguity among six equally scored Phase 1 candidates, which remained unresolved from the initial narrative alone.

To demonstrate the explainability of the final output, Listing~\ref{list:test-result-reasoning} shows a subset of the structured reasoning trace generated by the system for the top candidate, \emph{Hepatitis~A}.
\begin{lstlisting}[frame=single, breaklines=true, basicstyle=\footnotesize\ttfamily, aboveskip=1.5em, belowskip=0.3em, caption={Agent-generated reasoning trace for the refined diagnosis of Hepatitis A}, label={lst:test-result-reasoning}]
"rank": 1,
"disease": "hepatitis A",
"baseline_score": 0.861,
"followup_support": 0.095,
"followup_conflict": 0.083,
"final_score": 0.872,
"updated_confidence": "HIGH",
"reasoning": "Hepatitis A initially had a score of 0.861. Follow-up questions provided support by confirming the presence of jaundice (weight +0.733), but conflicted by denying ascites (weight -0.467), acholic stool (weight -0.733), and liver inflammation (weight -0.733). The net effect was an increase in the final score to 0.872, with updated confidence rated as HIGH"
\end{lstlisting}
\label{list:test-result-reasoning}

\subsection{Comparative analysis against the MedQDx baseline}

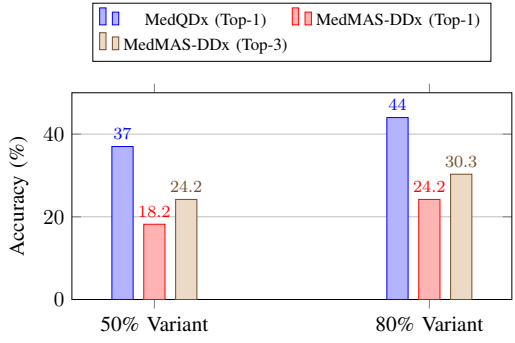
\begin{figure}
  \centering
  \begin{tikzpicture}[scale=0.8]
    \begin{axis}[
        ybar=5pt,
        bar width=10pt,
        width=\linewidth,
        height=5cm,
        enlarge x limits=0.3,
        ymin=0, ymax=50,
        ylabel={Accuracy (\%)},
        ylabel near ticks,
        symbolic x coords={50\% Variant, 80\% Variant},
        xtick=data,
        nodes near coords,
        nodes near coords style={font=\footnotesize},
        legend style={at={(0.5,1.15)}, anchor=south, legend columns=2, font=\footnotesize},
        ymajorgrids,
    ]
      \addplot coordinates {(50\% Variant,37.0) (80\% Variant,44.0)};
      \addplot coordinates {(50\% Variant,18.2) (80\% Variant,24.2)};
      \addplot coordinates {(50\% Variant,24.2) (80\% Variant,30.3)};
      \legend{MedQDx (Top-1), MedMAS-DDx (Top-1), MedMAS-DDx (Top-3)}
    \end{axis}
  \end{tikzpicture}
  \caption{Performance comparison: MedQDx vs MedMAS-DDx}
  \label{fig:medqdx_comparison}
\end{figure}

Figure~\ref{fig:medqdx_comparison} compares MedMAS-DDx against the MedQDx baseline (GPT-4.1) across both accuracy metrics and completeness levels.
MedQDx achieves higher zero-shot Top-1 accuracy ($44\%$ at $80\%$; $37\%$ at $50\%$), reflecting a different task formulation rather than better reasoning: as an open-vocabulary classifier, it is never penalized for missing labels at the benchmark's exact granularity, whereas MedMAS-DDx grounds each candidate in an explicit knowledge graph and refuses to generate a diagnosis with no ontology correspondent. This constraint enables greater explainability and traceability, as each prediction can be linked to structured evidence and a transparent reasoning process.

Figure~\ref{fig:medqdx_comparison} shows the same comparison at Top-3, the appropriate granularity for a differential diagnosis: the gap narrows to $30.3\%$ vs. $44\%$ at $80\%$ completeness, but less so at $50\%$ ($24.2\%$ vs. $37\%$), where Top-3 also falls below its own Phase~1 value - the same pattern discussed above.

The two systems also differ sharply in question budget: $2.4$ for MedQDx versus roughly ten for MedMAS-DDx, a direct consequence of confirming specific ontology-mapped symptoms rather than a single free-form hypothesis. Whether this cost is acceptable depends on the deployment context, discussed further in the Discussion.

\section{ Discussions and limitations}

In this section, we discuss the behavioral characteristics  and the possible limitations of the proposed system.

Our approach uniquely coordinates specialized agents via deterministic control logic rather than unconstrained communication. This architecture provides predictable execution, allowing validation of every intermediate decision. Consequently, the system suits safety-critical applications where reproducibility and auditability are essential.

The system leverages multi-turn questioning to progressively enrich the clinical picture. When initial input is sparse, the interaction naturally shifts toward targeted questions that collect discriminative evidence. Each turn refines the differential diagnosis by incorporating new responses into the ranking, enabling gradual convergence toward accurate conclusions. This highlights the strength of structured dialogue in improving diagnostic precision while maintaining transparent, traceable reasoning.

Overall, the contribution of interactive refinement is real but modest, and at $50\%$ completeness it is not even uniformly positive: Top-1 accuracy improves at both completeness levels (Fig.~\ref{fig:accuracy_phases}), but Top-3 accuracy improves only at $80\%$ completeness ($25.3\% \rightarrow 30.3\%$) and decreases slightly at $50\%$ ($26.3\% \rightarrow 24.2\%$), a pattern that also makes the narrowing of the gap with the MedQDx baseline uneven across variants. 

However, the case study (Section~\ref{sec:qualitative-case-study}) reveals a different dynamic. Case \#96 shifts one instance's Top-1 similarity from $0.340$ to $1.0$ by resolving ambiguity among six equally scored candidates via a single targeted confirmation. When initial ambiguity exists, refinement resolves it decisively. Conversely, if one diagnosis dominates or no candidate is sufficiently close to the reference label, refinement has no actionable uncertainty, yielding a zero measured contribution by construction.

The Top-3 drop at 50\% completeness arises because incomplete symptom information admits weak candidates into the Top-3. Follow-up evidence legitimately replaces these incidental candidates via the score update (Eq.~\eqref{eq:4}) with better-supported ones, occasionally reducing Top-3 accuracy. Aggregating cases with and without resolvable ambiguity inherently understates refinement's per-case impact; case-level trajectories provide a more faithful assessment of the follow-up loop.

Despite the advantages, several limitations should be acknowledged. First, the quality of the differential diagnosis depends on the completeness and accuracy of the underlying biomedical ontology and its symptom-disease associations. Missing, incomplete, or outdated knowledge cannot be incorporated into the reasoning process. 

Other key limitation in medical diagnosis tasks is the mismatch between ontology terminology and dataset terminology, which arises from the gap between formal, structured medical knowledge (ground truth) and the context-dependent, heterogeneous language used in clinical data as it is recorded in real-world practice. This gap limits the alignment of patient data with ontology concepts, affecting tasks such as symptoms-disease mapping or information extraction.
This limitation is partially mitigated by the extensive validation and widespread adoption of biomedical ontologies in research  ~\cite{zhang2019unsupervisedannotationphenotypicabnormalities},~\cite{Schriml2012-lk},~\cite{10.1093/nar/gkaa1043 } 
that explicitly leverage DOID or HPO, which provide a robust semantic foundation for clinical reasoning and
 as well as their successful application in ontology-based systems across various domains 
\cite{8116981_onto}, \cite{inbook}.

The knowledge base's limitations are exposed by two reference labels, \emph{Jaundice} and \emph{Vertigo}: while the Q4Dx benchmark treats them as diagnoses, SYMP classifies them as symptoms. Because MedMAS-DDx candidates must correspond to explicit disease concepts in the knowledge graph, the system cannot output either label as a diagnosis, regardless of reasoning quality. This stems directly from ontology coverage, not the reasoning pipeline.
For \emph{Jaundice}, this limitation is absolute: it is never proposed as a candidate disease. Instead, the system retrieves ontology-supported diseases featuring \emph{jaundice} (e.g., \emph{Hepatitis~E, D}) and queries it as follow-up evidence. Accuracy here remains zero until the ontology is explicitly extended.

Analyzing unresolved cases post-Phase 2 reveals that errors stem primarily from structured reasoning constraints rather than language understanding. Typical errors involve: (1) conceptual mismatches, where the benchmark treats symptoms as diagnoses; (2) knowledge graph sparsity, where exact symptom-disease associations are absent despite highly accurate symptom extraction (88.4\% mapping); (3) diagnostic granularity, where the system robustly identifies the broader disease family but struggles to isolate exact subtypes without specific laboratory results.

Consequently, employing a more advanced LLM would not resolve these errors, as its role is deliberately restricted to semantic translation. These unresolved cases highlight an intentional limitation of our neuro-symbolic design: constraining inference to a curated knowledge graph structurally prevents hallucinating statistically plausible but unsupported diagnoses. This trade-off caps benchmark performance to guarantee transparent, clinically safe decision-making.

Finally, the restricted interaction model limits follow-up responses to "yes/no/unsure." While supporting deterministic scoring and reproducibility, this fails to capture nuanced clinical information like symptom severity or temporal progression.

\section{Conclusion and Future Work}

Overall, the methodology forms a directed pipeline with a single feedback loop: free text is transformed into structured symptoms, generating candidate diseases consolidated into an initial ranking that either terminates or triggers targeted questioning. The follow-up answers are then reintegrated into the scoring stage. Throughout the pipeline, each component has a distinct role: language models perform interpretation and generation, the external knowledge source provides symptom-disease associations, and deterministic code handles ranking, evaluation, and control flow. A sparse initial description simply produces a weaker initial ranking and shifts more work onto the questioning stage, allowing the inquiry depth to adapt to initial uncertainty. This consistency keeps the process transparent at all stages.

The evaluation of the proposed method showed that interactive refinement improved global Top-1 accuracy from 18.7\% to 21.2\% and Top-3 accuracy from 25.8\% to 27.3\%. Notably, targeted follow-up questions recovered diagnostic confidence to levels comparable to the zero-shot semantic-similarity baseline using substantially richer initial narratives.

Despite these promising results, several development opportunities remain. The system requires validation on large-scale, real-world datasets (e.g., MIMIC-IV~\cite{PhysioNet-mimiciv-3.1, Stark2025-xj} or DDXPlus~\cite{tchango2022ddxplusnewdatasetautomatic}), whose clinical notes, laboratory results, and diverse patient cases better reflect real-world complexity, addressing the current benchmark's limited size. Future efforts will extend ontology coverage, enable richer patient interactions while preserving determinism, and conduct prospective clinical evaluations. Finally, integrating adaptive weighting and probabilistic evidence could further enhance accuracy of the ranking without compromising diagnostic transparency.

\section{Acknowledgement} 

The authors acknowledge Georgiana Copil (Adobe Romania) for her valuable guidance and coordination. 

\bibliographystyle{IEEEtran}

\bibliography{sample_bib} 
\end{document}